\documentclass[runningheads]{llncs}
\usepackage{xcolor}
\usepackage{graphicx}
\usepackage{rotating} 

\usepackage{amsfonts,amsmath}

\begin{document}
\title{Learning Graph Representations}
%
%
\author{Rucha Bhalchandra Joshi\orcidID{0000-0003-1214-7985} \and
Subhankar Mishra\orcidID{0000-0002-9910-7291}}
\authorrunning{R. Joshi and S. Mishra}
%
\institute{National Institute of Science Education and Research, Bhubaneswar - 752050, India \\
Homi Bhabha National Institute, Anushaktinagar, Mumbai - 400094, India\\
\email{\{rucha.joshi, smishra\}@niser.ac.in}} 

\maketitle              
\begin{abstract}

Social and information networks are gaining huge popularity recently due to their various applications. Knowledge representation through graphs in the form of nodes and edges should preserve as many characteristics of the original data as possible. Some of the interesting and useful applications on these graphs are graph classification, node classification, link prediction, etc. The Graph Neural Networks have evolved over the last few years.
Graph Neural Networks (GNNs) are efficient ways to get insight into large and dynamic graph datasets capturing relationships among billions of entities also known as knowledge graphs.

In this paper, we discuss the graph convolutional neural networks graph autoencoders and spatio-temporal graph neural networks. The representations of the graph in lower dimensions can be learned using these methods. The representations in lower dimensions can be used further for downstream machine learning tasks.

\keywords{Graph Neural Networks \and
Graph Embeddings \and
Graph Representation Learning}
\end{abstract}

\section{Introduction}
The complex networks like social networks, citation networks etc. can very well be represented in form of graphs. The questions related to these graphs can be addressed using neural networks. However, feeding the graphs as input to the neural networks needs to be done cleverly. This is due to the vast and complex nature of the graphs. The nodes of the graph need to be represented in lower dimensional vector form so that it can easily be given as input to the neural network to address the problems. These representations should characterize the information contained in the graphs. They should capture the structural as well as the feature information contained in the graph. The representations in lower dimensional space can be then fed to a neural network to address the tasks on the graphs. These representations are necessary in order to reduce to some extent the overhead caused due to the large size of the network.

\begin{figure}
    \centering
    \includegraphics[width=0.9\textwidth]{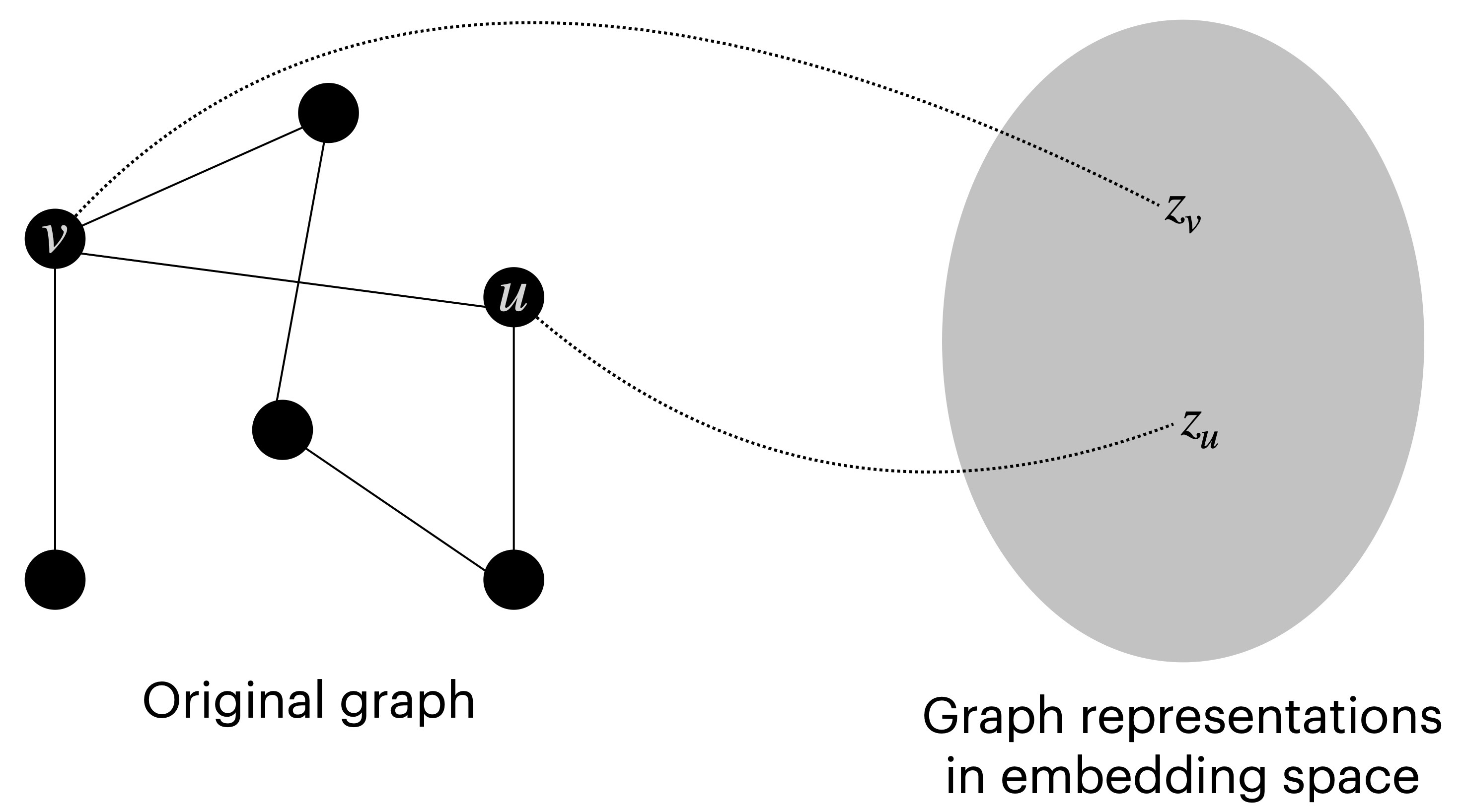}
    \caption{Representing nodes in embedding space. Similar nodes in original graph must be similar in embedding space.}
    \label{fig:graph-to-emb}
\end{figure}

These representations in lower dimensional space are also called the embeddings. Several shallow embedding techniques like DeepWalk \cite{perozzi2014deepwalk}, Node2vec \cite{grover2016node2vec} were earlier introduced to generate node representations. They considered only the vertex set and the adjacency matrix of the graph in order to generate the embeddins. An encoding function and a similarity function that measures the similarity between similar nodes in the embedding space need to be defined. The parameters are optimized based on this similarity function. The shortcoming of such methods are, however, the large number of parameters to be optimized and also that these methods did not consider the feature information to generate the node embeddings. To overcome these shortcomings, the deep graph neural networks are used. There have been several approaches to produce the representations for nodes which considered the node features in addition to the structure of the node neighborhood. Initial methods are from spectral graph theory, which mostly require matrix factorization. To improve on such methods the spatial based methods were introduced. These methods consider information diffusion and message passing. The neighborhood feature information is aggregated to generate the representation of a particular node. Figure \ref{fig:graph-to-emb} is the portrayal of nodes in original graph mapped to their representations in embedding space. 

In this paper we review the popular graph neural network methods that are primarily based on neighborhood aggregation. The graph neural networks were introduced by \cite{gori}. Then there were several improvements on the same in terms of how the neighborhood aggregation is done. The neighborhood aggregation is looked at as the convolution over the node in the graph. The spectral graph theory has the notion of convolutions on the graph where a graph signal, which is feature vector of a node in this case, convolved in the Fourier domain. This idea have been used in the spectral based methods whereas the spatial based methods broadly rest on the concept of neighborhood aggregation. The graphs autoencoders aims at reconstructing the original adjacency matrix. The architecture includes an encoder and a decoder. We have discussed the graph auto-encoding techniques in this paper. The spatio-temporal graph neural networks are designed to work with the data that can be easily modeled using spatio-temporal graphs. The graph changes over the period of time, depending on the system that is being modeled. The graph neural network used to generate the embedding for such data structure should consider the spatial as well as the temporal dependencies in the graph. We discuss some of the spatio-temporal graph neural network methods in this paper.

The rest of the paper is organized as follows. In section \ref{sec:prelim} we discuss the preliminaries. We discuss spectral and non-spectral graph convolutional methods in section \ref{sec:cgnn}, autoencoding techniques in \ref{sec:gae} and spatio-temporal graph neural networks in section \ref{sec:stgnn}. Section \ref{sec:discussions} is discussions and section \ref{sec:future-dir} discusses the future directions.  We conclude the paper with section \ref{sec:conclusion}.

\section{Preliminaries}
\label{sec:prelim}
A \emph{graph} is represented as $G = (V, E)$ where $V$ is a set of vertices or nodes and $E$ is the set of edges. Let $v_i \in V$ be the $i$th node in the vertex set $V$. The edge connecting nodes $v_i$ and $v_j$ is represented as $e_{ij} = (v_i, v_j)$. $n$ is the number of nodes in the graph and $m$ is the number of edges. $\mathbf{A}$ is the adjacency matrix of the graph. The dimension of $\mathbf{A}$ is $n \times n$. The entry $\mathbf{A}_{ij} = 1$ if $e_{ij} \in E$ . The neighbourhood $N(v)$ of a node $v$ is defined as $N(v) = \{u \in V | (v, u) \in E\}$.

The feature vector corresponding to node $v$ is represented as $\mathbf{x}_v \in \mathbb{R}^{d}$ where $d$ is the size of node feature vector. The matrix of feature vectors of nodes in a graph is $\mathbf{X} \in \mathbb{R}^{n \times d}$. The hidden representation corresponding to node $v$ is given as $\mathbf{h}_v^k$ in $k$th layer of the Graph Neural Network. The final representation corresponding to a node in the embedding space is denoted as $\mathbf{z}_v$

Table \ref{tab_notations} summarizes the frequently used notations with their descriptions.

\begin{table}
\centering
\caption{Notations}
\label{tab_notations}
\begin{tabular}{|l|l|}
\hline
Notation &  Description\\
\hline
$G$     & A graph\\
$V$     & The set of nodes\\
$E$     & The set of edges\\
$v_i$   & A vertex in $V$\\
$e_{i,j}$ & An edge in $E$ connecting vertices $v_i$ and $v_j$\\
$n$     & Number of nodes\\
$m$     & Number of edges\\
$\mathbf{A}$    & The adjacency matrix of graph $G$\\
$\mathbf{D}$    & The diagonal matrix of node degrees of graph $G$\\
$N(v)$  & The neighbourhood of $v$\\
$\mathbf{x}_v$  & Feature vector of node $v$\\
$\mathbf{X}$    & The feature matrix of graph $G$\\
$d$     & The dimension of the node representation in previous layer\\
$d'$    & The dimension of the node representation generated in current layer\\
$\mathbf{h}_v^k$& Node $v$'s representation in $k$th hidden layer\\
$\mathbf{H}^k$  & Matrix of node representations in $k$th hidden layer\\
$\mathbf{z}_v$  & Node $v$'s representation in the embedding space\\
$\mathbf{Z}$    & Matrix of node representations in the embedding space\\
$\mathbf{L}$    & Laplacian of a matrix\\
$\mathcal{L}$   & Normalized Laplacian of a matrix\\
$\mathbf{U}$    & Matrix of eigenvalues\\
$\mathbf{\Lambda}$  & Matrix of eigenvectors\\
$\mathbf{\lambda}_i$    & $i$th eigenvectors\\
$\theta$    & Learnebale convolutional kernel\\
$\mathbf{\Theta}$   & Diagonal matrix of learnebale convolutional kernels\\
$\sigma$    & Non-linear activation function\\
$\mathbf{W}_k, \mathbf{B}_k$    & Learneable parameters for $k$th layer\\
$AGG$   & Some aggregation function\\
$MEAN$  & Element-wise mean aggregator\\
$\gamma$& Element-wise pooling operator\\
$LSTM$  & Long short term memory aggregator\\
$\pi$   & Permutation function\\
$\alpha_{ij}$   & Attention coefficient between $i$ and $j$\\
$LeakyRELU$ & Leaky rectified linear unit activation\\
$\epsilon$  & A learneable parameter to determine a node's contribution to the next layer update\\
$p$     & Distribution of input data\\
$q$     & Estimation of distribution of latent variable\\
$\mathbb{E}$& Expectation\\
$KL(\cdot||\cdot)$  & KL divergence between two distributions\\
$\mathcal{N}(\mu, \sigma)$   & Normal distribution with mean $\mu$ and standard deviation $\sigma$\\
$\star_G$   & Diffusion convolution operator\\
$r^{(t)}$     & Reset gate vector\\
$u^{(t)}$     & Update gate vector\\
$C^{(t)}$     & Candidate gate vector\\
\hline
\end{tabular}
\end{table}

\section{Convolutional Graph Neural Networks}
\label{sec:cgnn}
The convolutional graph neural networks draw motivations from the conventional convolutional neural networks(CNNs). These graph neural networks can be divided into two categories based on the principle they work on. The \emph{spectral based} convolutional graph neural networks and the \emph{spatial based} convolutional graph neural networks. The early spectral based methods essentially makes use of the adjacency matrix and degree matrix of the graph and perform the convolution operation in the Fourier domain. The convolution filter is applied to the graph signal in the Fourier domain, and then transformed back to the graph domain. The limitation of these kind of convolutional methods is that these are inherently transductive and hence need the entire graph for the computation. In the transductive setting, the model is specific to the graph it is trained on, and may not be extendable to the graph that has not been seen by the model. These methods also do not use the shared parameters. In contradiction to transductive methods, the inductive graph neural network methods are such that the parameters can be used to generate the embeddings for unseen graphs.

The graph structure is given by the adjacency matrix. Although a particular adjacency matrix is not the only way to represent the graph. There could be multiple adjacency matrices corresponding to a particular graph. Hence, the representation in lower dimensional space should be specific to the graph and not to the adjacency matrix. Irrespective of any permutation of the adjacency matrix of a graph, the graph neural network should give the same representation for a particular task.
It is essential that the function generating the representations should either be invariant or equivariant, depending on whether the task at hand is graph level or node level. For the graph level task like graph classification, the graph neural network function needs to be permutation invariant as described in the following equation:

\begin{equation}
    f\left( Q\mathbf{A}Q^T, \mathbf{X}\right) = f(\mathbf{A}, \mathbf{X})
\end{equation}

where, $Q$ is a permutation matrix, $\mathbf{A}$ is adjacency matrix of the graph, $\mathbf{X}$ is the feature matrix corresponding to the graph and $f$ denotes the function approximated by the neural network. 

For node level tasks like node classification, the graph neural network function needs to be permutation equivariant as given in the equation below:

\begin{equation}
    f\left( Q\mathbf{A}Q^T, \mathbf{X}\right) = Q f(\mathbf{A}, \mathbf{X})
\end{equation}

\subsection{Spectral Based}

The spectral based methods perform operations on the graph by making use of the adjacency matrix of the graph, Laplacian matrix of the graph etc. The Laplacian of the graph is the measure of smoothness of it. The \emph{non-normalized graph Laplacian} is defined as $\mathbf{L} = \mathbf{D} - \mathbf{A}$ where $\mathbf{D}$ is the diagonal degree matrix of the graph $G$ defined as $\mathbf{D}_{ii} = \sum_{j} \mathbf{A}_{ij}$. The \emph{normalized graph Laplacian} is $\mathcal{L} = \mathbf{I}_n - \mathbf{D}^{-1/2}\mathbf{A}\mathbf{D}^{-1/2}$. $\mathcal{L}$ is a real symmetric positive semidefinite matrix. It has a complete set of orthogonal eigenvectors with associated real positive eigenvalues. The matrix can be factorized as $\mathcal{L} = \mathbf{U}\mathbf{\Lambda}\mathbf{U}^T$ where $\mathbf{U} = [u_0, u_1, ... ,u_{n-1}] \in \mathbb{R}^{n\times n}$ is a matrix of eigenvectors arranged according to eigenvalues and $\mathbf{\Lambda}$ is a diagonal matrix with eigenvalues $\{\lambda_0, \lambda_1, ... ,\lambda_{n-1}\}$ as diagonal elements.

The graph signal is a function $\mathbf{x}$ that maps vertices to $\mathbb{R}$. 
The \emph{graph Fourier transform} of a given graph signal $\mathbf{x}$ is defined as $\hat{f}(\mathbf{x})=\mathbf{U}^{T} \mathbf{x}$ and the \emph{inverse graph Fourier transform} is given by $f(\hat{\mathbf{x}}) = \mathbf{U}\mathbf{\hat{x}}$.
This working of the convolutional graph neural networks is show in figure 

\begin{figure}
    \centering
    \includegraphics[width=\textwidth]{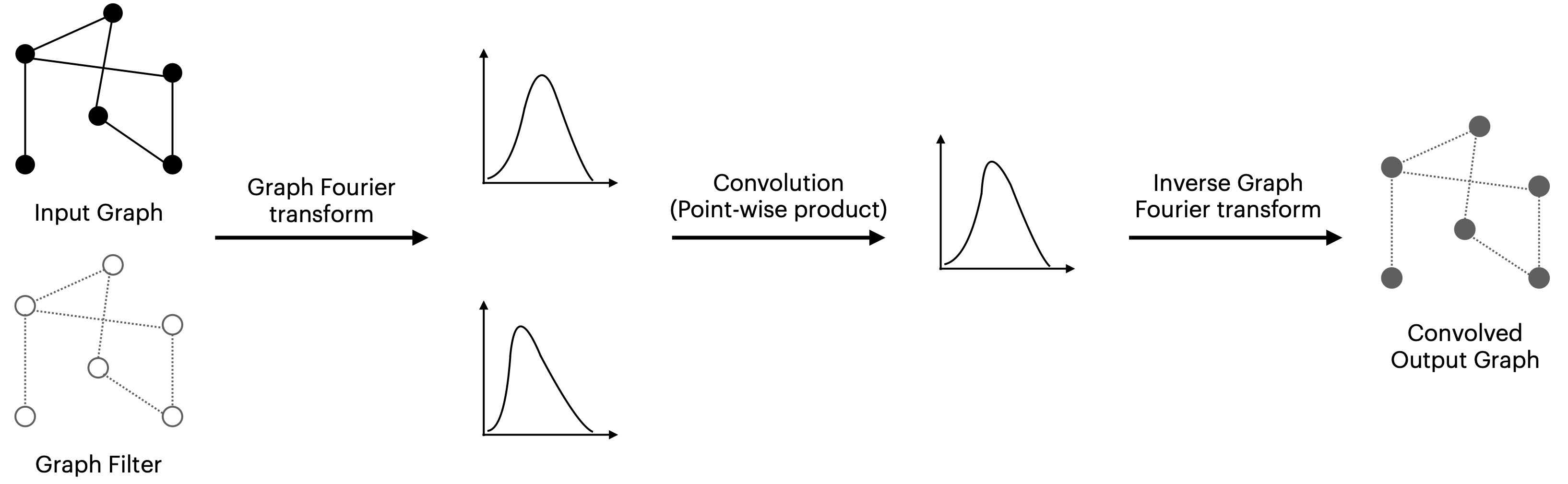}
    \caption{Working of spectral based graph convolutional methods. The graph and the kernel is convolved once they are transformed into the Fourier Domain. The result of the convolution is transformed back to as the convolved graph by applying the inverse Fourier transform.}
    \label{fig:spectra-conv}
\end{figure}

The \emph{convolution} operation in the spectral based is methods is defined by applying a filter which is a diagonal matrix $\mathbf{g}_{\theta} = diag(\theta)$, where $\theta \in \mathbb{R}^n$ to the Fourier transform of the signal as follows:

\begin{equation}
    \mathbf{g}_\theta * \mathbf{x} = \mathbf{U}\mathbf{g}_\theta\mathbf{U}^T\mathbf{x}
\end{equation}

\emph{Spectral graph convolutional networks} \cite{bruna2013spectral} considers the signal with multiple $K$ channels and hence $f_k$ filters in layer $k$ for $k = 1,...,K$. The input signal to each of the layer is $\mathbf{H}^{k-1}_i \in \mathbb{R}^{n\times f_{k-1}}$ and the output signal generated is $\mathbf{H}^{k}_i \in \mathbb{R}^{n\times f_{k}}$. The convolution operation is defined as:
\begin{equation}
    \mathbf{H}_j^{k+1} = \sigma(\mathbf{U} \sum_{i=1}^{f_{k-1}}\mathbf{\Theta}^k_{i, j}\mathbf{U}^T\mathbf{H}_j^k) \quad (j=1,...,f_{k-1})
\end{equation}
where $\mathbf{\Theta}^k_{i,j}$ is diagonal matrix of learnable parameters. This operation costs $O(n^3)$.

\emph{ChebNet} \cite{defferrard2016convolutional} approximates the filter $\mathbf{g}_\theta$ as truncated Chebyshev's polynomials up to $K$ degree of the polynomial of the diagonal matrix of the eigenvalues $\mathbf{\Lambda}$. The Chebyshev's polynomials are recursively defined as $T_0(x)=1, T_1(x)=x, T_{k+1}(x)=2xT_{k}(x)-T_{k-1}(x)$. The convolutional filter is parameterized as:
\begin{equation}
    \mathbf{g}_\theta (\mathbf{\Lambda}) = \sum_{k=0}^{K}\theta_k T_{k}(\mathbf{\hat{\Lambda}})
\end{equation}
where $\theta \in \mathbb{R}^K$ is vector consisting Chebyshev's coefficients, $T_k(\hat{\Lambda})$ is defined at $\hat{\Lambda} = \frac{2\Lambda}{\lambda_{max}}-\mathbf{I}_n$. The convolution operation on the graph signal $\mathbf{x}$ using the above filter is hence defined as:
\begin{equation}
    \mathbf{g}_\theta * \mathbf{x} = \sum_{k=0}^{K}\theta_k T_{k}(\mathbf{\hat{\mathbf{L}}})\mathbf{x}
\end{equation}
where $\hat{\mathbf{L}} = \frac{2\mathcal{L}}{\lambda_{max}} - \mathbf{I}_n$ and $T_k(\hat{\mathbf{L}}) = \mathbf{U}T_k(\hat{\Lambda})\mathbf{U}^T$. ChebNet reduces the complexity of the filtering operation from $O(n^3)$ to $O(Km)$.

\emph{CayleyNets} \cite{levie2018cayleynets} define the convolution filter as follows:
\begin{equation}
    \mathbf{x} * \mathbf{g}_\theta = c_0\mathbf{x} + 2Re\{ \sum_{j=0}^{r} c_j (h\mathcal{L} - i\mathbf{I})^{j} (h\mathcal{L} + i\mathbf{I})^{-j} \mathbf{x}\}
\end{equation}

where $\mathcal{L}$ is normalized graph Laplacian matrix, $Re(.)$ gives the real part of the Cayley's polynomial, $c$ and $h$ are learnable parameters. 

The filters proposed by \emph{Graph Convolutional Networks}(GCN) \cite{kipf2016semi} convolve over the graph and limits the $K$ value from ChebNet to 2. Having the larger value of $K$ may over smoothen the information in the node, by aggregating too much infromation from the neighborhood nodes. The convolution operation for GCN is:
\begin{equation}
    \label{eq:gcn}
    \mathbf{g}_\theta * \mathbf{x} = \theta_0 \mathbf{x} + \theta_1 \mathbf{D}^{-\frac{1}{2}} \mathbf{A} \mathbf{D}^{-\frac{1}{2}} \mathbf{x}
\end{equation}
where $\theta_0 $ and $\theta_1$ are learnable parameters. Equation \ref{eq:gcn} can be written in matrix form as follows:
\begin{equation}
    \mathbf{H} = \tilde{\mathbf{D}}^{\frac{1}{2}} \tilde{\mathbf{A}} \tilde{\mathbf{D}}^{\frac{1}{2}} \mathbf{X} \mathbf{\Theta}
\end{equation}
where $\tilde{\mathbf{A}} = \mathbf{A} + \mathbf{I}_n$ and $\tilde{\mathbf{D}}_{ii} = \sum_{j} \tilde{\mathbf{A}}_ij$. The identity matrix is added to adjacency matrix in order to include the contribution from the node itself.

Dual Graph Convolutional Networks(DGCN) introduced by \cite{zhuang2018dual} applies two graph convolutions on same inputs in order to capture the local and the global consistencies. It uses Positive Pointwise Mutual Information(PPMI) matrix for encoding the information.

\subsection{Non Spectral Based}

The graph neural networks that are not spectral based primarily use the principle of message passing. The node feature information is shared across the neighborhood of the node. The methods differ in the way they apply the aggregations over the neighborhood. 

\begin{figure}
    \centering
    \includegraphics[width=0.35\textwidth]{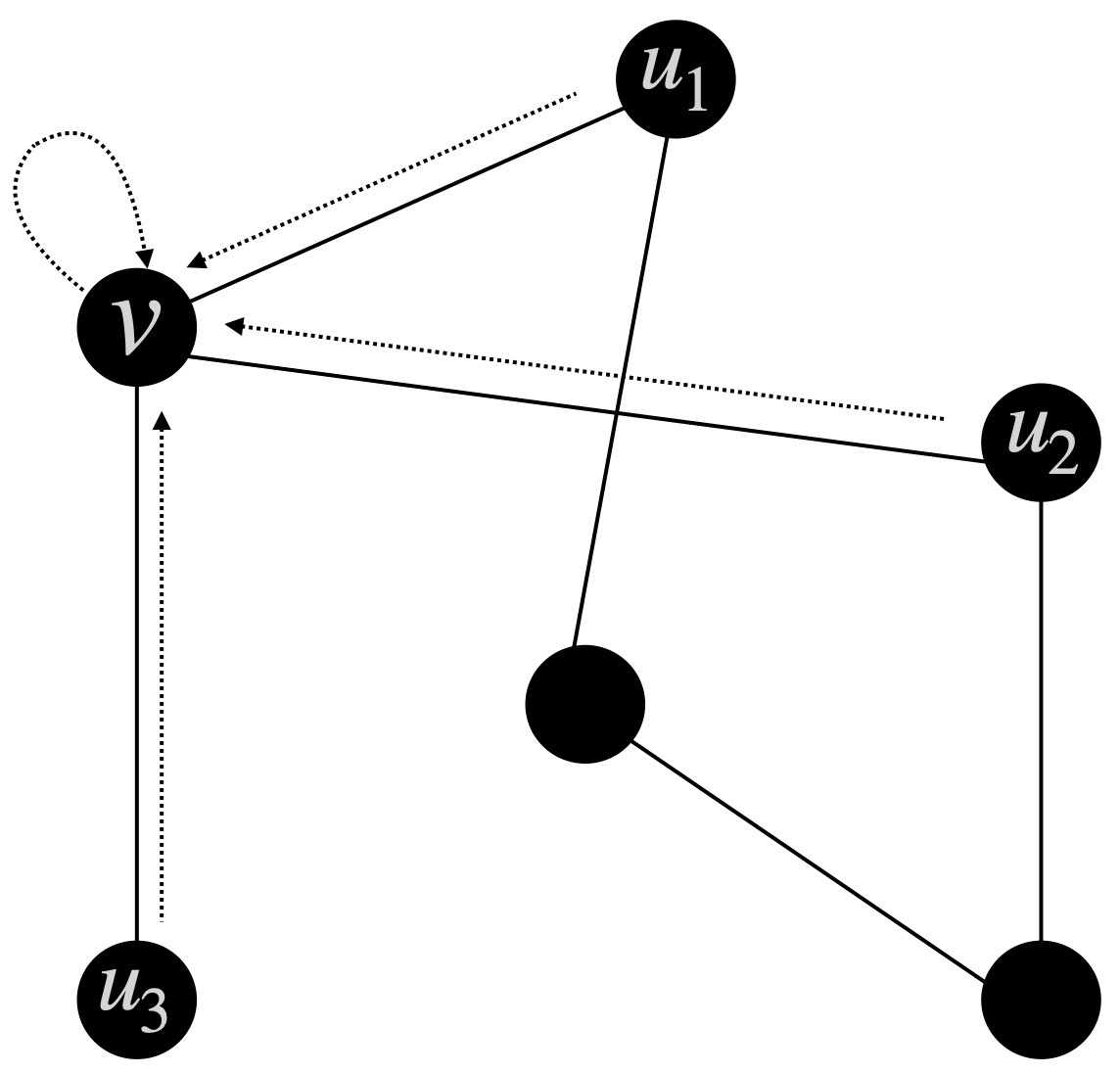}
    \caption{Neighborhood aggregations. To generate the representation of vertex (e.g. $v$), the spatial based methods consider the contributions from the neighborhood nodes($u_1, u_2, u_3$ in this example). The methods specify how the neighborhood information is aggregated.}
    \label{fig:neighbors}
\end{figure}

NN4G \cite{micheli2009neural} is a constructive feedforward neural network. The neighborhood information is summed up which is why it causes the hidden states to have different scales. 

The GCN model aggregates neighborhood features and reduces the impact of high degree neighboring nodes. Equation \ref{eq:gcn} can also be viewed as follows for each of the nodes $v$ in $k$th layer: 
\begin{equation}
    \label{eq:gcn_nodewise}
    \mathbf{h}_v^k = \sigma \left( \mathbf{W}_k \sum_{u \in \mathcal{N}(v)\bigcup v} \frac{\mathbf{h}_u^{k-1}}{\sqrt{|\mathcal{N}(u)||\mathcal{N}(v)|}}  \right)
\end{equation}
The contribution of the high degree neighbors is lowered in this manner. The GCN model works in trasductive manner for fixed graphs. Many of the earlier approaches optimize the node embeddings using matrix factorization based objective functions.

An inductive approach called GraphSAGE was introduced by \cite{hamilton2017inductive}. GraphSAGE samples a fixed size neighborhood of a node and applies the aggregators on those. Unlike the previous transductive approaches, GraphSAGE does not generate embeddings unique to a particular graph. The trained model can be applied to a new similar graph. The embedding for a node $v$ is computed as follows:
\begin{equation}
    \mathbf{h}_v^k = \sigma \left(  \mathbf{W}_k \cdot CONCAT \left( \mathbf{h}_v^{k-1}, AGG \left( \{ \mathbf{h}_u^{k-1}, \forall u \in \mathcal{N}(v) \}\right) \right) \right)
\end{equation}
where, $AGG$ is aggregation function applied on the neighborhood of a node. The aggregated neighborhood is concatenated with the current node's previous layer representation.
The three aggregators examined in \cite{hamilton2017inductive} are as follows:

The \emph{mean aggregator} which takes element-wise mean of the neighborhood representations of a node as given in equation \ref{eq:mean_agg}.

\begin{equation}
    \label{eq:mean_agg}
    AGG = MEAN\left( \left\{ \mathbf{h}_v^{k-1} \right\} \cup \left\{ \mathbf{h}_u^{k-1}, \forall u \in \mathcal{N}(v) \right\} \right)
\end{equation}

The \emph{pooling aggregator} is given in equation \ref{eq:pooling_agg}, where a neural network is applied to the representations of neighborhood nodes and then element-wise max-pool operation($\gamma$) is applied.

\begin{equation}
    \label{eq:pooling_agg}
    AGG = \gamma \left( \left\{ \sigma' \left( \mathbf{W}_{pool} \mathbf{h}_u^{k-1} + b \right), \forall u \in \mathcal{N}(v) \right\} \right)
\end{equation}

The third aggregator is \emph{Long Short Term Memory(LSTM) aggregator}. The LSTM architecture depends on the order of the input hence the aggregator is applied to a random permutation of the neighbors. It is given in equation \ref{eq:lstm_agg}.

\begin{equation}
    \label{eq:lstm_agg}
    AGG = LSTM \left( \left[ \mathbf{h}_u^{k-1}, \forall u \in \pi\left( \mathcal{N}(v) \right) \right] \right)
\end{equation}

\begin{figure}
    \centering
    \includegraphics[width=0.37\textwidth]{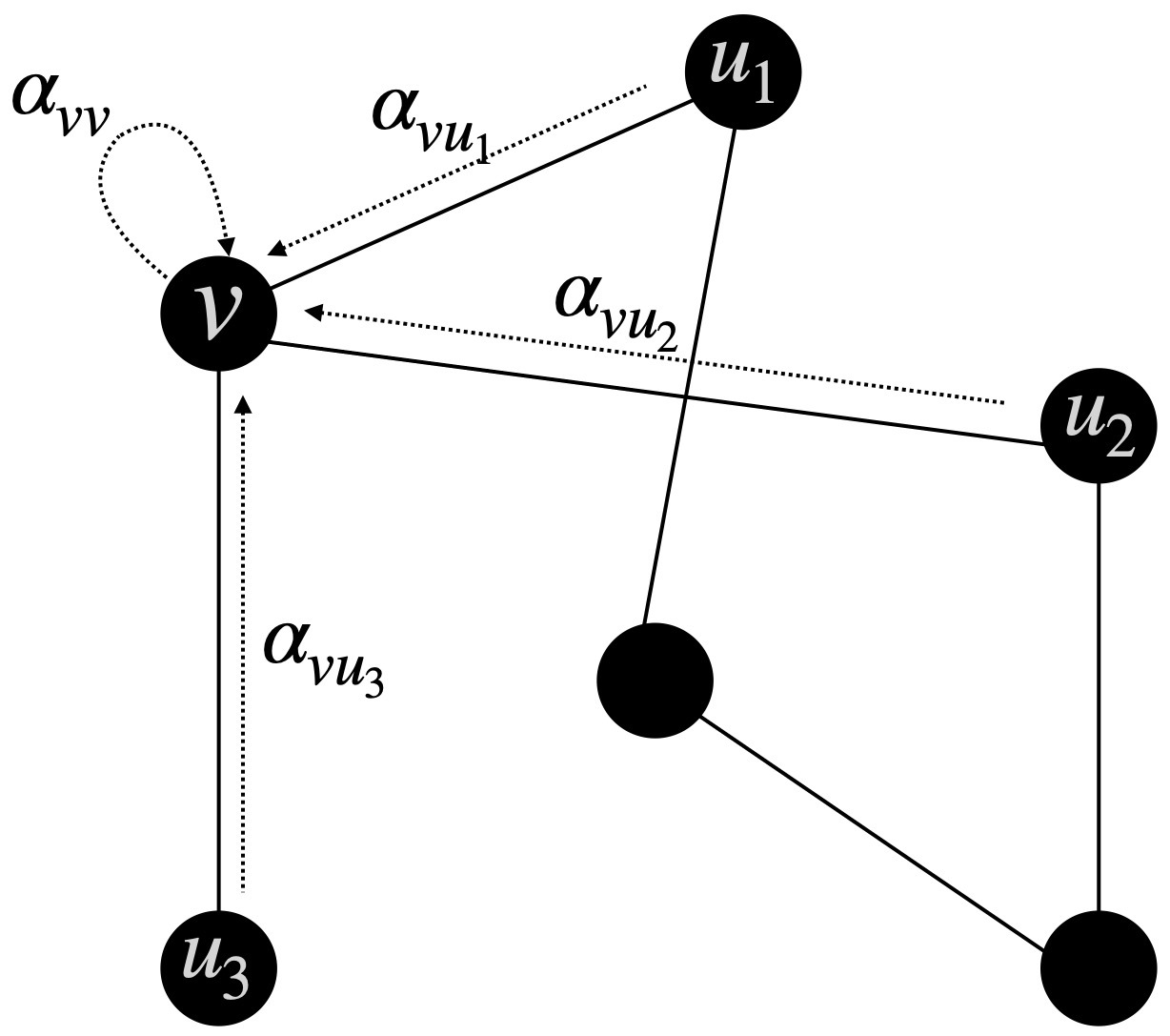}
    \caption{Attention coefficients. Each of the neighbors may not contribute equally to the representation of a node, hence the attention coefficient is calculated in Graph Attention Networks. In this example, the attention coefficients are shown as $\alpha_{vu_i}$ where $u_i$ are neighbors of $v$.}
    \label{fig:attention}
\end{figure}

The Graph Attention Networks \cite{velickovic2018graph} introduced attention based method to generate the node representation by taking attending over the neighborhood of the node. The attention coefficient of a node and each of its neighbor is computed as given in equation \ref{eq:attention_coef} and a graphical representation is shown in figure \ref{fig:attention}.

\begin{equation}
\label{eq:attention_coef}
    \alpha_{ui} = \frac{exp\left( LeakyReLU \left( \mathbf{a}^T \left[ \mathbf{W} \mathbf{h}_u \vert \mathbf{W} \mathbf{h}_i \right] \right) \right)}{\sum_{j \in \mathcal{N}(v)} exp\left( LeakyReLU \left( \mathbf{a}^T \left[ \mathbf{W} \mathbf{h}_v \vert \mathbf{W} \mathbf{h}_j \right] \right) \right)}
\end{equation}

where we compute importance of node $i$'s features to node $v$'s representation. It is normalized using softmax over the neighborhood. The attention mechanism used here is Multi-Layered Perceptron applied to the concatenation of the linearly transformed representations of the node $v$ and node $i$. The attention coefficient is further used to take the weighted linear combination of the neighborhood representations of the node in order to form next layer representation for the current node. A non-linear function $\sigma $ is applied to the linear combination as given by equation \ref{eq:gat}.
\begin{equation}
    \label{eq:gat}
    \mathbf{h}_v^{k} = \sigma \left( \sum_{i \in \mathcal{N}(v) \cup v} \alpha_{vi} \mathbf{W}\mathbf{h}_i^{k-1} \right)
\end{equation}

To stabilize the learning process, the \emph{multi-head} attention is used. The attention coefficients are calculated by applying different attention mechanisms in different attention heads. The representations generated using each of the attention mechanisms are either concatenated as in equation \ref{eq:gat_concat} which increases the dimensions of the new representation to $\mathbb{R}^{L \times d'}$. We consider that there are $L$ attention heads. 

\begin{equation}
    \label{eq:gat_concat}
    \mathbf{h}_v^{k} = \Big{\Vert}_{l=1}^L \sigma \left( \sum_{i \in \mathcal{N}(v)\bigcup v} \alpha_{vi}^{k,l} \mathbf{W}_k^l \mathbf{h}_i^{k-1} \right)
\end{equation}

The above equation generates the representation with dimensions $L$ times more. In order to stabilize the learning as well as generate representation of size $d'$, the $L$ representations are averaged as in equation \ref{eq:gat_avg}.

\begin{equation}
    \label{eq:gat_avg}
    \mathbf{h}_v^{k} = \sigma \left( \frac{1}{L} \sum_{l=1}^L \sum_{i \in \mathcal{N}(v)\bigcup v} \alpha_{vi}^{k,l} \mathbf{W}_k^l \mathbf{h}_i^{k-1} \right)
\end{equation}

The non-linearity is applied after taking the average of the neighborhood representation from each of the attention heads.

Graph Isomorphism Networks \cite{xu2018how} is a framework that generalizes the Weisfeiler-Lehman (WL) \cite{weisfeiler1968reduction} isomorphism test. The node feature representation update function is given as follows:

\begin{equation}
    \mathbf{h}_v^k = MLP^{(k)} \left( \left( 1 + \epsilon^{(k)} \right) \cdot \mathbf{h}_v^{k-1} + \sum_{u \in \mathcal{N}(v)}\mathbf{h}_u^{k-1} \right)
\end{equation}
where $\epsilon$ is a learneable parameter.
The graph neural network is said to be powerful if the aggregation function it uses is injective, that is if it maps the non isomorphic graphs to different representations. In the previous models, max pooling or mean pooling were used to do the neighborhood aggregation. These schemes fail to fully capture the structural dissimilarity between two dissimilar graphs. GIN uses the sum pooling and manages to map nodes with different structure to different node representations.

In some corner cases GIN fails to produce the different representations for nodes with different rooted subtree structure. These cases are resolved using Relational Pooling \cite{murphy2019relational} architecture.

\begin{table}
    \centering
    \caption{Various methods discussed in the paper.}
    \begin{tabular}{|p{0.17\linewidth} | p{0.2\linewidth} | p{0.63\linewidth}|}
    \hline
     Category    &  Method  &  Update \\
     \hline
     Spectral based &  Spectral Graph Convolutional Networks \cite{bruna2013spectral} & $\mathbf{H}_j^{k+1} = \sigma(\mathbf{U} \sum_{i=1}^{f_{k-1}}\mathbf{\Theta}^k_{i, j}\mathbf{U}^T\mathbf{H}_j^k) \quad (j=1,...,f_{k-1})$ \\[3ex]
     \cline{2-3}
     & ChebNet \cite{defferrard2016convolutional} & $\mathbf{H} = \sum_{k=0}^K T_k(\hat{\mathbf{L}})\mathbf{X}\mathbf{\Theta}_k$ \\[2ex]
     \cline{2-3}
     & GCN \cite{kipf2016semi} & $\mathbf{H} = \tilde{\mathbf{D}}^{\frac{1}{2}} \tilde{A} \tilde{\mathbf{D}}^{\frac{1}{2}} \mathbf{X} \mathbf{\Theta}$ \\[2ex]
     \hline
     Non spectral based & GraphSAGE \cite{hamilton2017inductive}& $\mathbf{h}_v^k = \sigma \left( \left[ \mathbf{W}_k \cdot AGG \left( \{ \mathbf{h}_u^{k-1}, \forall u \in \mathcal{N}(v) \}\right), \mathbf{B}_k \mathbf{h}_v^{k-1} \right] \right)$\\[2ex]
     \cline{2-3}
     & Graph Attention Networks \cite{velickovic2018graph}& $\mathbf{h}_v^{k} = \sigma \left( \sum_{i \in \mathcal{N}(v) \cup v} \alpha_{vi} \mathbf{W}\mathbf{h}_i^{k-1} \right)$ \\[2ex]
     \cline{2-3}
     & Graph Isomorphism Networks \cite{xu2018how}& $\mathbf{h}_v^k = MLP^{(k)} \left( \left( 1 + \epsilon^{(k)} \right) \cdot \mathbf{h}_v^{k-1} + \sum_{u \in \mathcal{N}(v)}\mathbf{h}_u^{k-1} \right)$ \\[2ex]
     \hline
     Graph Autoencoders & GAE \cite{kipf2016variational} & $ \mathbf{Z}= GCN \left( \mathbf{X}, \mathbf{A} \right) \quad \textnormal{where} \quad \hat{\mathbf{A}} = sigmoid \left( \mathbf{ZZ}^T \right)$ \\[2ex]
     \cline{2-3}
     & Linear GAE \cite{salha2019keep} \cite{salha2020simple}& $\mathbf{Z} = \tilde{\mathbf{A}}\mathbf{W} \quad and \quad \hat{\mathbf{A}} = sigmoid (\mathbf{Z}\mathbf{Z}^T)$ \\[2ex]
     \hline
     Spatio-temporal Graph Neural Networks  & DCRNN \cite{li2017diffusion} & $r^{(t)} = \sigma \left( \mathbf{W}_r \star_G \left[ \mathbf{X}^{(t)}, \mathbf{H}^{(t-1)} \right]+b_r\right)$ \par
     $u^{(t)} = \sigma \left( \mathbf{W}_u \star_G \left[ \mathbf{X}^{(t)}, \mathbf{H}^{(t-1)} \right]+b_u\right)$ \par
     $C^{(t)} = \tanh \left( \mathbf{W}_c \star_G \left[ \mathbf{X}^{(t)}, \left(r^{(t)} \odot \mathbf{H}^{(t-1)} \right) \right]+b_c\right)$ \par
     $\mathbf{H}^{(t)} = u^{(t)} \odot \mathbf{H}^{(t-1)} + \left( 1-u^{(t)} \right) \odot C^{(t)}$ \\[2ex]
     \cline{2-3}
     & STGCN \cite{yu2018spatio} & $\mathbf{H}^{(l+1)} = \mathbf{\Gamma}^l_1 *_\tau ReLU \left( \Theta^l * \left( \mathbf{\Gamma}_0^l *_\tau \mathbf{H}^l \right) \right)$\\[2ex]
     \hline
    \end{tabular}
    
    \label{tab:methods}
\end{table}

\section{Graph Autoencoders}
\label{sec:gae}
\begin{figure}
    \centering
    \includegraphics[width=0.85\textwidth]{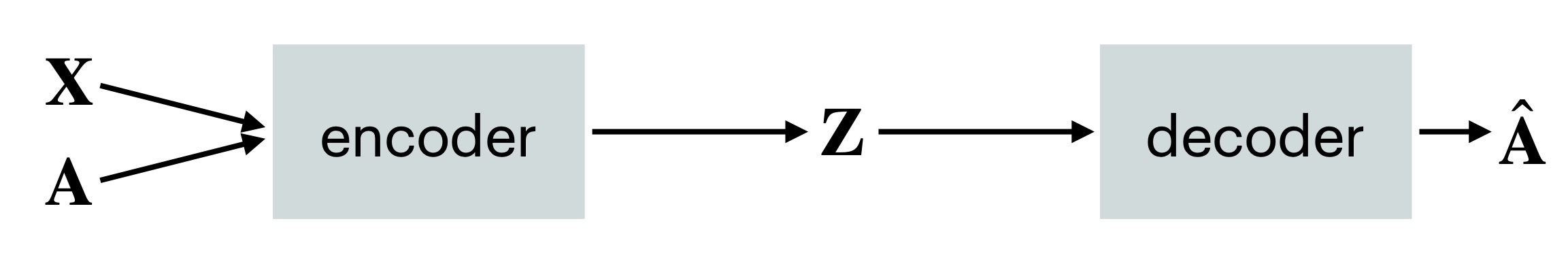}
    \caption{Graph Autoencoder(GAE). The $\mathbf{X}$ and $\mathbf{A}$ are the feature matrix and the adjacency matrix of the graph respectively. $\mathbf{Z}$ is the matrix of latent representations and $\hat{\mathbf{A}}$ is the reconstructed adjacency matrix.}
    \label{fig:gae}
\end{figure}
The graph autoencoders(GAE) consists of two parts, encoder and the decoder. The encoder produces the latent representations corresponding to the nodes of the graph and the decoder part of a GAE aims at reconstructing the original adjacency matrix of the graph. The GAE is shown in the figure \ref{fig:gae}.

\begin{figure}
    \centering
    \includegraphics[width=\textwidth]{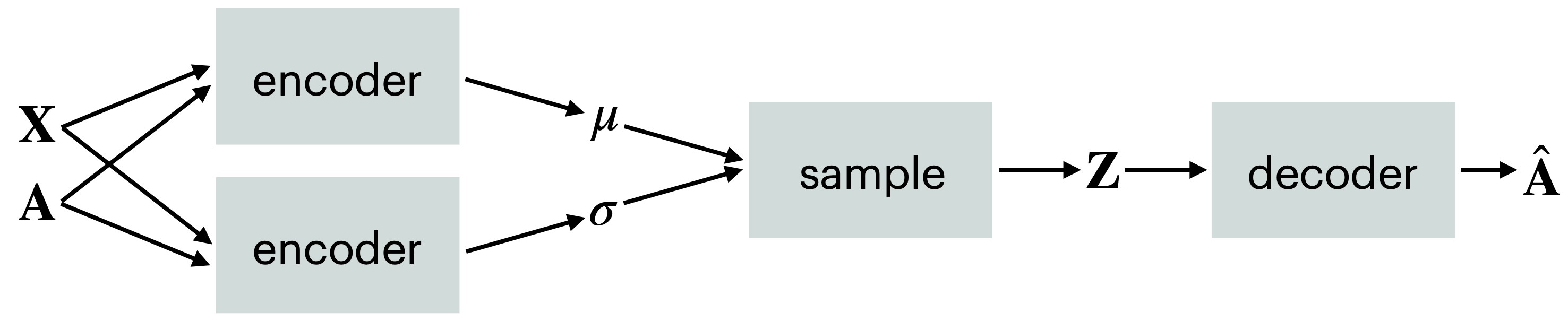}
    \caption{Graph Variational Autoencoder(GAE). The mean $\mu$ and variance $\sigma$ of the the input is calculated by two respective encoders in order to know the distribution. }
    \label{fig:gvae}
\end{figure}

The variational graph autoencoders learns the latent representations in a probabilistic way. \cite{kipf2016variational} introduced Graph variational auto-encoders(VGAE). The conventional working of a GVAE is shown in the figure \ref{fig:gvae}. The VGAE uses two layered GCN \cite{kipf2016semi} as the inference model and the inner product of the representations as the generative model. The inference model or the encoder of VGAE that consists of GCN layers is as follows:

\begin{equation}
    q\left(\mathbf{A}|\mathbf{Z}\right) = \prod_{i=1}^N q\left(\mathbf{z}_i|\mathbf{X}, \mathbf{A}\right), \quad \textnormal{where} \quad q\left(\mathbf{z}_i | \mathbf{X}, \mathbf{A}\right) = \mathcal{N}\left(\mathbf{z}_i | \mu_i, diag\left(\sigma^2_i\right)\right)
\end{equation}
where $\mu$ and $\sigma$ are calculated using GCNs. The first layer of their GCNs share the weights. The two layers of the GCN work as per equation \ref{eq:gcn_nodewise}, which can be written as $GCN(\mathbf{X}, \mathbf{A}) = \tilde{\mathbf{A}} \textnormal{ReLU}\left( \tilde{\mathbf{A}} \mathbf{XW}_0 \right)W_1$, where $W_0$ and $W_1$ are weight matrices corresponding to first and second layer of the GCN and the matrix $\tilde{\mathbf{A}} = \mathbf{D}^{-\frac{1}{2}} \mathbf{A} \mathbf{D}^{-\frac{1}{2}}$. The generative model or the decoder for the GVAE is given based on the similarities of the latent space variables, which are generated using the \lq reparameterization trick\rq  \cite{kingma2013auto}. This is necessary as the back-propagation cannot be done when $q\left(\mathbf{z}_i | \mathbf{X}, \mathbf{A}\right)$ is sampled, since it cannot be differentiated. The generative model or the decoder is as follows:

\begin{equation}
    p\left( \mathbf{A} | \mathbf{Z} \right) = \prod_{i=1}^N \prod_{j=1}^N p\left( A_{ij} | \mathbf{z}_i, \mathbf{z}_j \right), \quad \textnormal{where} \quad p\left( A_{ij} = 1 | \mathbf{z}_i, \mathbf{z}_j \right) = sigmoid\left( \mathbf{z}_i^T \mathbf{z}_j \right)
\end{equation}

The objective function for GVAE is 
\begin{equation}
    \mathcal{L} = \mathbb{E}_{q(\mathbf{Z}|\mathbf{X}, \mathbf{A})} \left[ \textnormal{log} \ p(\mathbf{A}|\mathbf{Z}) \right] - \textnormal{KL}[q(\mathbf{Z}|\mathbf{X}, \mathbf{A}) || p(\mathbf{Z})]
\end{equation}
where, the first term gives the expectation of the likelihood of generating the adjacency matrix $A$ when the latent variables are sampled from the distribution $q(\mathbf{Z}|\mathbf{X}, \mathbf{A})$. The second term in the equation calculates the Kullback-Leibler divergence between the two distributions $q(\mathbf{Z}|\mathbf{X}, \mathbf{A})$ and $p(\mathbf{Z})$. 

The non-probabilistic version of the Graph Autoencoders(GAE) consists of only the two layer GCN as encoder and the inner product of the matrices of latent variables as decoder that reconstructs the adjacency matrix $\hat{\mathbf{A}}$. This is given in the following equations.
\begin{equation}
    \hat{\mathbf{A}} = sigmoid \left( \mathbf{ZZ}^T \right) \quad \textnormal{where} \quad \mathbf{Z}= GCN \left( \mathbf{X}, \mathbf{A} \right)
\end{equation}
The reconstruction loss in GAE is minimized as follows:
\begin{equation}
    \mathcal{L} = \mathbb{E}_{q(\mathbf{Z}|\mathbf{X},\mathbf{A})}\left[ log \ p(\hat{\mathbf{A}}|\mathbf{Z}) \right]
\end{equation}

The shortcoming of the GVAE is that it primarily preserves only the topological structure of the graph. The Adversarially Regulated Graph Autoencoders(ARGA) \cite{pan2018adversarially} encodes the contents of the nodes along with the topological structure of graph. This model makes use of adversarial regularization. The latent representations are matched with the prior distribution so that the discriminator can discriminate the latent variable $\mathbf{z}_i$ if is from the encoder or from the prior distribution. Adversarially Regulated Variational Graph Autoencoders(ARVGA) are also discussed in the paper where it makes use of the VGAE instead of a GAE.

The autoencoding techniques that we have discussed makes use of GCNs as an encoder. \cite{salha2019keep}, \cite{salha2020simple} recently proposed to use a linear model as an encoder instead of the GCN. This model is called Linear Graph Auto Encoders(LGAE). The encoder and decoder of LGEA is as follows:
\begin{equation}
    \mathbf{Z} = \tilde{\mathbf{A}}\mathbf{W} \quad and \quad \hat{\mathbf{A}} = sigmoid (\mathbf{Z}\mathbf{Z}^T)
\end{equation}
where $\mathbf{W}$ is the weight matrix corresponding to the linear encoder. The encoder linearly maps the adjacency matrix to the latent space. 

The encoder of the linear graph variational autoencoders(LGVAE) outputs the distribution by giving $\mu$ and $\sigma$ as follows:

\begin{equation}
    \mu = \tilde{\mathbf{A}}\mathbf{W}_\mu \quad \textnormal{and} \quad \textnormal{log}\ \sigma = \tilde{\mathbf{A}}\mathbf{W}_\sigma \quad \textnormal{then} \quad \mathbf{z}_i \sim \mathcal{N}(\mu_i, diag(\sigma^2_i))
\end{equation}
Except the encoder, rest of the model is similar to that of the GVAE as discussed above.

\section{Spatio-temporal Graph Neural Networks}
\label{sec:stgnn}

The spatio-temporal systems have dependency on both time and space. The dependency can be observed in a form of changes in the structure or the features over the period of time. If the systems can be well modeled using a graph, we need to also incorporate the changes that occur in this graph. These type of graphs are dominated by temporal in addition to spatial and feature information as in earlier cases. The neural networks that would be used to find the embeddings of these graph, hence, need to consider the changes in the graph happening over the period of time. We can broadly categorize these neural networks as spatio-temporal graph neural networks(STGNNs).

The typical choice for handling the spatial dependencies is using some neural network that produces the embedding considering the structural and feature information of the graph. A typical example of this could be using a Graph Convolutional Network \cite{kipf2016semi}. Whereas in order to handle the temporal information, a suitable variation of a recurrent neural network may be used. While this is a very broad way of finding embeddings using STGNNs, various other combinations can be made based on the need of the system. 

The Graph Convolutional Recurrent Neural Network \cite{seo2018structured} proposed two architectures. Both the architectures use the LSTM in order to model the temporal dependency. They differ from each other in the method that is used to model the spatial dependency. One of the two uses the Convolutional Neural Networks over graph, while the other one uses the Graph Convolutions. 

\begin{sidewaysfigure}
    \centering
    \includegraphics[width=\textwidth]{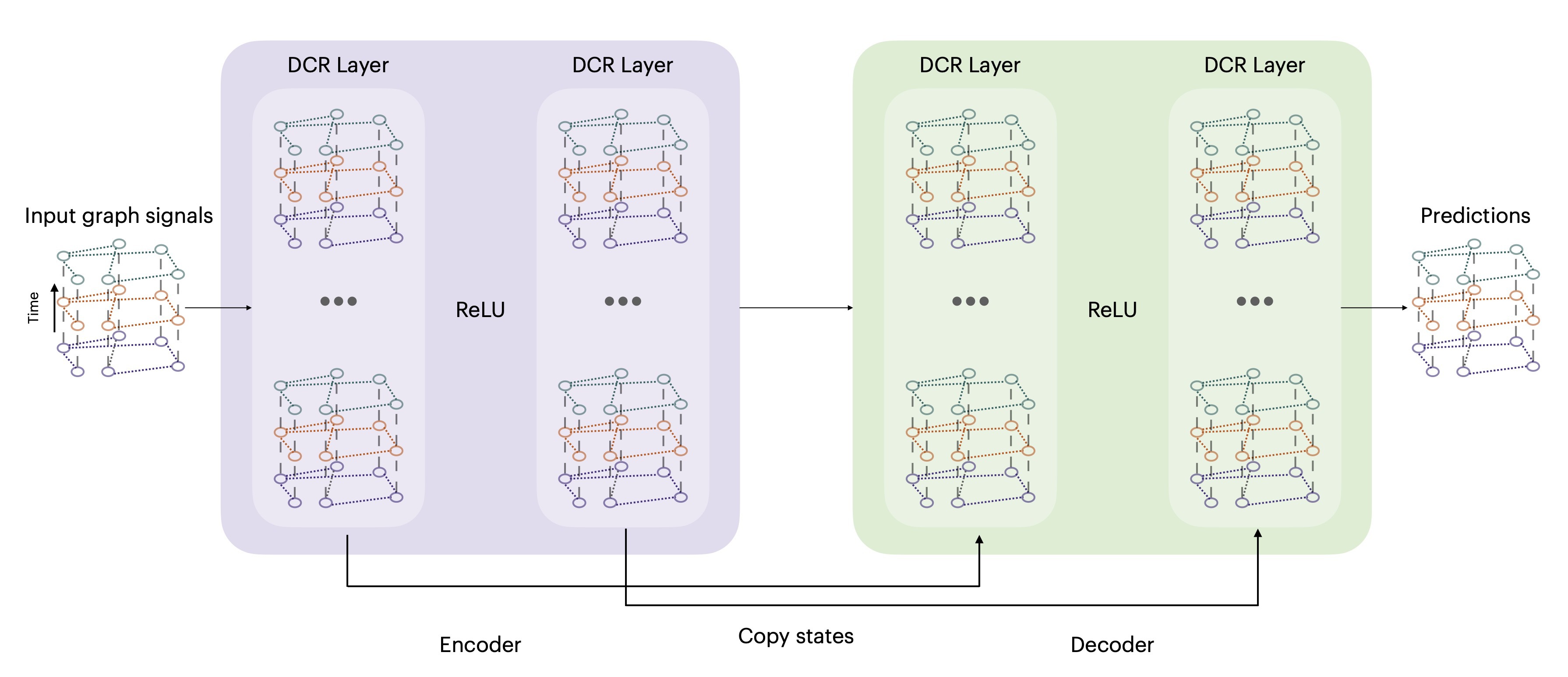}
    \caption{Diffusion Convolutional Recurrent Neural Networks \cite{li2017diffusion}}
    \label{fig:dcrnn}
\end{sidewaysfigure}

The Diffusion Convolution Recurrent Neural Network(DCRNN) \cite{li2018diffusion} uses the diffusion convolution to model the spatial dependency while the temporal dependency is modeled using a Gated Recurrent Unit(GRU) \cite{cho2014learning}.The diffusion is defined as follows:
\begin{equation}
    \mathbf{X} \star_G \mathbf{W} = \sum_{k=0}^{K} \left( \mathbf{W}_O {\left( \mathbf{D}_O^{-1} \mathbf{A} \right)}^k + \mathbf{W}_I {\left( \mathbf{D}_I^{-1} \mathbf{A}^T \right)}^k \right)\mathbf{X}
\end{equation}
where, $\star_G$ is a diffusion convolution operation over graph $G$. The diffusion operation is applied $K$ times. $\mathbf{W}_O$ and $\mathbf{W}_I$ are learneable parameters for bidirectional diffusion. $\mathbf{D}_O$ is the out-degree matrix and $\mathbf{D}_I$ is the in-degree matrix.

The diffusion convolution layer is defined as follows:
\begin{equation}
    \mathbf{H} = \sigma \left( \mathbf{X} \star_G \mathbf{W} \right)
\end{equation}
where, $\mathbf{W} \in \mathbb{R}^{d'\times d}$, $\mathbf{H} \in \mathbb{R}^{n\times d}$ and $\mathbf{X} \in \mathbb{R}^{n\times d'}$. The diffusion convolution can be applied in directed as well as undirected setting. Note that the spectral graph convolution is similar diffusion convolution when applied in undirected graphs. 

The temporal dependency is captured using GRU as follows:
\begin{subequations}
    \begin{align}
    r^{(t)} &= \sigma \left( \mathbf{W}_r \star_G \left[ \mathbf{X}^{(t)}, \mathbf{H}^{(t-1)} \right]+b_r\right) \\
    u^{(t)} &= \sigma \left( \mathbf{W}_u \star_G \left[ \mathbf{X}^{(t)}, \mathbf{H}^{(t-1)} \right]+b_u\right) \\
    C^{(t)} &= \tanh \left( \mathbf{W}_c \star_G \left[ \mathbf{X}^{(t)}, \left(r^{(t)} \odot \mathbf{H}^{(t-1)} \right) \right]+b_c\right) \\
    \mathbf{H}^{(t)} &= u^{(t)} \odot \mathbf{H}^{(t-1)} + \left( 1-u^{(t)} \right) \odot C^{(t)}
    \end{align}
\end{subequations}
The terms used in STGNNs are listed in the table \ref{tab:dgcrn}. 
The encoder-decoder architectures is used in DCRNNs to make the predictions based on the data. The encoder generates a fixed length embeddings for the data which is then passed to a decoder that makes the predictions.

\begin{table}[]
    \centering
    \caption{Diffusion Convolution Recurrent Neural Network Terminology}
    \begin{tabular}{l l}
    $r^{(t)}$     & Reset gate vector\\
    $u^{(t)}$     & Update gate vector\\
    $C^{(t)}$     & Candidate gate vector\\
    $\mathbf{H}^{(t)}$  & Output vector\\
    $\mathbf{W}, \mathbf{b}$     & Parameter matrix and vector\\
    \end{tabular}
    \label{tab:dgcrn}
\end{table}

The Dynamic Diffusion Convolutional Recurrent Neural Network(D-DCRNN) \cite{mallick2020dynamic} is similar to DCRNN, as it has an encoder decoder architecture with both of them having the convolutional and the recurrent layers. The adjacency matrix in the D-DCRNN is dynamic in nature which is computed from the current state of the data.

\begin{figure}
    \centering
    \includegraphics[width=0.35\textwidth]{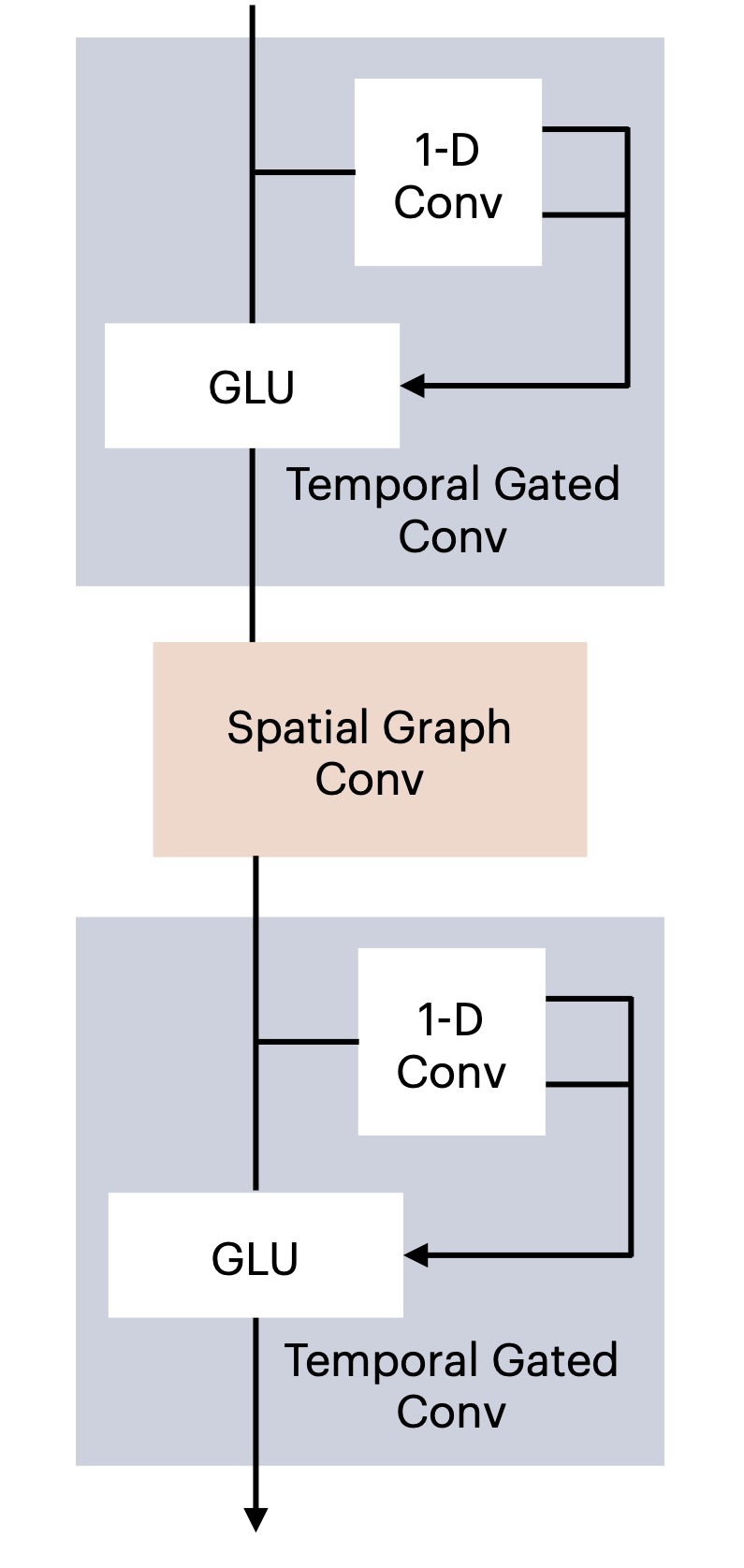}
    \caption{The spatio-temporal block in STGCN \cite{yu2018spatio} that sandwiches a spatial convolutional block between two temporal gated convolutional blocks.}
    \label{fig:stgcn}
\end{figure}

Most of the STGNN architectures have been designed by keeping the traffic networks in mind. The recurrent networks are slow with respect to aspects such as complex gate mechanisms, dynamic changes etc. In order to overcome these, the architecture called Spatio-temporal Graph Convolutional Networks(STGCN), proposed in \cite{yu2018spatio}, makes use of convolutions. The STGCN architecture as shown in figure \ref{fig:stgcn} basically has two blocks of spatio-temporal convolutions followed by the output layer. Each of the spatio-temporal block sandwiches a spatial convolutional block between to temporal gated convolutional blocks as shown in the equation \ref{eq:stgcn}. Unlike the previously discussed architectures, STGCN does not have the encoder-decoder architecture, but it still incorporates the bottleneck strategy by making use of 64 filters for temporal gated convolutional blocks and 16 filters for Spatial graph convolution block. In each of the temporal gated convolutional block, residual connection is used. Gated Linear Unit(GLU) acts as an activation in these blocks. 

\begin{equation}
\label{eq:stgcn}
    \mathbf{H}^{(l+1)} = \mathbf{\Gamma}^l_1 *_\tau ReLU \left( \Theta^l * \left( \mathbf{\Gamma}_0^l *_\tau \mathbf{H}^l \right) \right)
\end{equation}
where $\mathbf{\Gamma}^l_0$ and $\mathbf{\Gamma}^l_1$ are the upper and lower temporal kernels in the spatio-temporal block $l$, $\mathbf{\Theta}$ is the spectral graph convolution kernel and $ReLU$ is the rectified linear unit.

\section{Discussion}
\label{sec:discussions}

While the previous approaches have been transductive in nature, the GraphSAGE was the approach introduced as inductive one. The inductive approaches can generate the embeddings for unseen networks using the model trained on similar graphs. Most of the approaches before GraphSAGE work in trasductive setting, that is, they cannot generalize to similar new data. Though they can be modified to work for new similar graphs, but the modification is computationally expensive. Graph attention networks do work in inductive manner. The results shown by GAT are better when compared with GraphSAGE. The weightage in form of attention coefficient corresponding to each pair of the neighbors, determines the contribution of the neighbor in the embedding. The graph structures that cannot be distinguished by the GNNs like GCN, GraphSAGE, are indentified in GIN paper. The GIN architecture is as powerful as WL-isomorphism test in distinguishing the graph structures.

\begin{table}[]
    \centering
    \caption{Summary of features exhibited by the methods discussed in this paper.}
    \begin{tabular}{|p{0.40\linewidth}|c|c|c|c|c|}
        \hline
        Method & Spectral & Spatial & Temporal & Transductive & Inductive\\
        \hline
        Spectral Graph Convolutional Networks \cite{bruna2013spectral} & $\bullet$ & $\circ$ & $\circ$ & $\bullet$ & $\circ$ \\
        ChebNet \cite{defferrard2016convolutional} & $\bullet$ & $\circ$ & $\circ$ & $\bullet$ & $\circ$ \\
        GCN \cite{kipf2016semi} & $\bullet$ & $\circ$ & $\circ$ & $\bullet$ & $\circ$ \\
        GraphSAGE \cite{hamilton2017inductive} & $\circ$ & $\bullet$ & $\circ$ & $\circ$ & $\bullet$ \\
        Graph Attention Networks \cite{velickovic2018graph} & $\circ$ & $\bullet$ & $\circ$ & $\circ$ & $\bullet$ \\
        Graph Isomorphism Networks \cite{xu2018how} & $\circ$ & $\bullet$ & $\circ$ & $\circ$ & $\bullet$ \\
        GAE \cite{kipf2016variational} & $\circ$ & $\bullet$ & $\circ$ & $\circ$ & $\bullet$ \\
        Linear GAE \cite{salha2019keep} \cite{salha2020simple} \cite{kipf2016semi} & $\circ$ & $\bullet$ & $\circ$ & $\circ$ & $\bullet$ \\
        DCRNN \cite{li2017diffusion} & $\circ$ & $\bullet$ & $\bullet$ & $\circ$ & $\bullet$ \\
        STGCN \cite{yu2018spatio} & $\circ$ & $\bullet$ & $\bullet$ & $\circ$ & $\bullet$ \\
        \hline
    \end{tabular}
    \label{tab:summary}
\end{table}

During the training of DCRNNs, the final states of encoder are passed on to the decoder. which then generates the predictions given the ground truth. During the testing time, the preditions replace the ground truth, which leads to variation in the distribution. In order to overcome this, the actual ground truth and predictions are given to the decoder with some probability $\epsilon_i$ and $1-\epsilon_i$ respectively. While these methods rely on the recurrent neural networks, STGCN does not employ it since the RNNs iterations are slower on the traffic data for which the model was essentially designed. To overcome all the shortcomings of using recurrent networks, the STGCN implements the convolutional networks across the temporal component. Though the models are especially designed for traffic forecasting, they can be extended for other systems that can be modeled as spatio-temporal graphs.

\section{Future Directions}
\label{sec:future-dir}
The existing graph representation learning techniques have their own strengths about producing representations. However, considering the scope we suggest the following directions for future work. Although there have been some work \cite{luo2020dynamic}, \cite{iiyama2020DistanceWeightedGN} in these directions, there is still a lot of significant work to be done. 
\begin{itemize}
    \item First, as these techniques work primarily for static graphs, whereas many real life systems modeled as graphs are dynamic in nature. Considering the dynamicity of the graphs, it is important to learn the representations of the dynamic graphs. 
    \item Second, multi-layered graphs' representations may need a different approach than the existing ones. 
    \item Thirdly, when there is diversity in the characteristics of the nodes in the graph, that is, the nodes exhibit heterogeneous behavior, the representations need to capture the variations in the working of the nodes. 
\end{itemize}  

\begin{figure}
    \centering
    \includegraphics[width=0.75\textwidth]{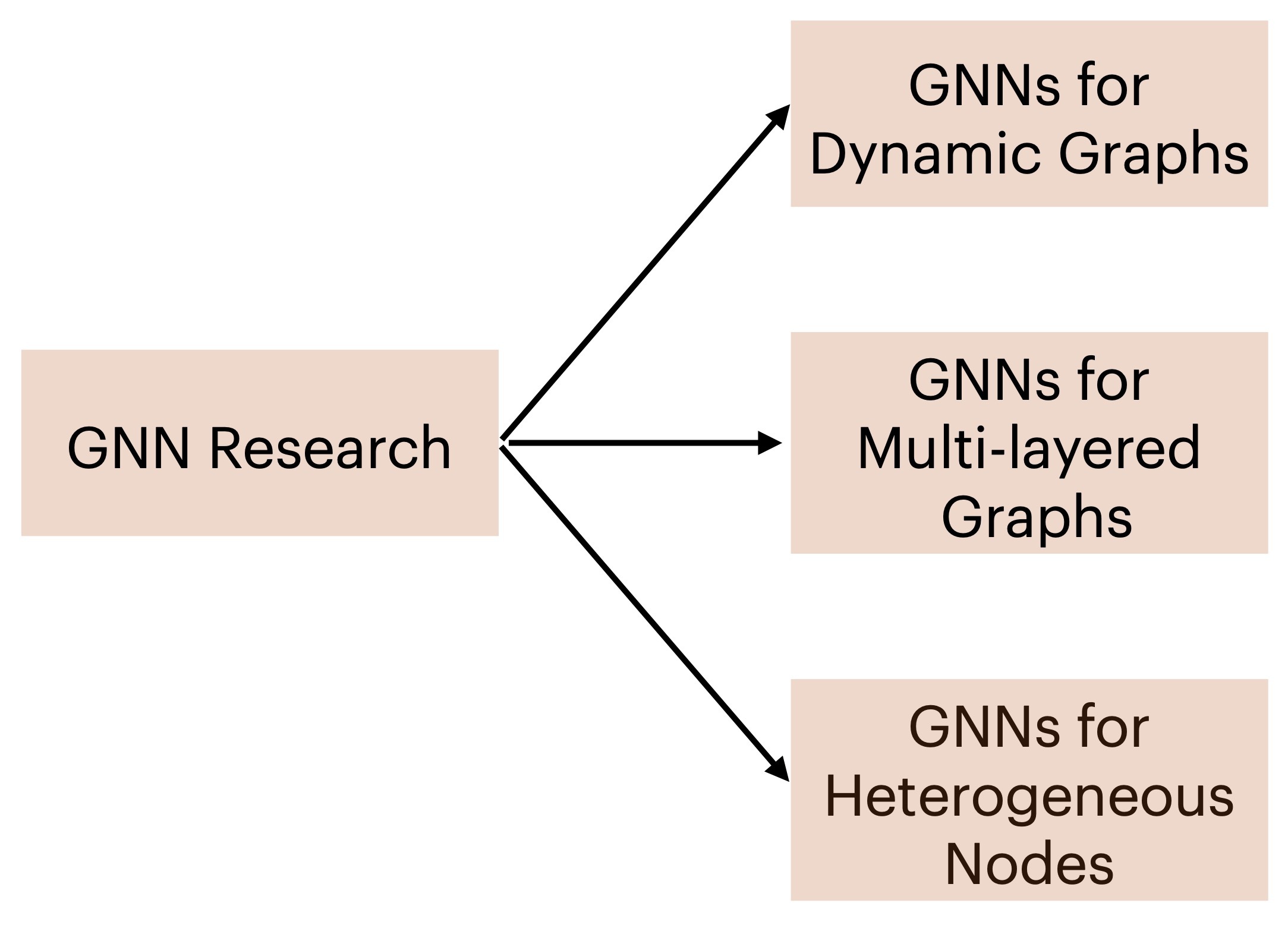}
    \caption{Future research directions}
    \label{fig:future-dir}
\end{figure}

\section{Conclusion}
\label{sec:conclusion}
In this paper, we have reviewed popular graph representation learning methods. These have been broadly categorized into spectral based methods and non-spectral based methods. The spectral based methods consider the spectrum of the graph in order to perform operations such as convolutions on it. The non spectral based methods makes use of the information from neighborhood of nodes. Each of the methods may have their own way of defining how the neighborhood information is aggregated. We also discussed graph autoencoders, which learn the latent representations for the graphs and based on which they reconstructs the graph adjacency matrix back. The variational autoencoders are also discussed, which are probabilistic models, which estimates the distribution of the data, from which the latent representation is chosen and the adjacency matrix is reconstructed. These autoencoders are helpful in the tasks such as link predictions, where it predicts the probability of any two nodes having an edge connecting them. We have also discussed some of the spatio-temporal graph neural networks that are used for the data that is modeled as spatio-tempoal graphs. The STGNNs use neural networks to include the spatial as well as the temporal dependencies in the graph. 

We further gave the future direction of research in this area. Real world examples that are modeled as dynamic graphs, can to be addressed using graph neural networks, which is an open area for research. Graph Neural Networks for multi-layered graphs and the graphs where different nodes have different physics behind their working are types of graphs other than static graphs, which need to be addressed using graph neural networks.

%
%
%
 \bibliographystyle{splncs04}
 \bibliography{ref}
\end{document}